\documentclass[letterpaper, 10 pt, conference]{ieeeconf}  

\usepackage{amsmath}
\usepackage{amssymb} 

\renewcommand{\QED}{\QEDopen}

\usepackage[labelformat=simple]{subcaption}
\DeclareCaptionLabelSeparator{periodspace}{.\quad}
\usepackage{tikz}
\usetikzlibrary{arrows,matrix,positioning,patterns}
\usetikzlibrary{calc}
\usepackage{pgfplots} 
\usetikzlibrary{plotmarks}

\usepackage[utf8]{inputenc}
\usepackage{pgfplots}
\usepgfplotslibrary{groupplots}

\usepackage{url}
\usepackage{multirow}
\usepackage{colortbl}

\usepackage[ruled, vlined, linesnumbered]{algorithm2e}

\DeclareSymbolFont{bbold}{U}{bbold}{m}{n}
\DeclareSymbolFontAlphabet{\mathbbold}{bbold}

\usepackage[labelformat=simple]{subcaption}
\DeclareCaptionLabelSeparator{periodspace}{.\quad}
\usepackage{listings}
\usepackage{xcolor}
\usepackage{cite}

\usepackage{wrapfig}

\IEEEoverridecommandlockouts                              
\title{\huge \bf
Automating Vehicles by Deep Reinforcement Learning\\ using Task Separation with Hill Climbing}

\author{Mogens Graf Plessen
\thanks{This work was mostly done while MGP was with IMT School for Advanced Studies Lucca, Piazza S. Francesco 19, 55100 Lucca, Italy, \texttt{mogens.plessen@alumni.imtlucca.it}.}
}

\renewcommand{\baselinestretch}{0.90}
\begin{document}

\maketitle
\thispagestyle{empty}
\pagestyle{empty}

\begin{abstract}
Within the context of autonomous driving a model-based reinforcement learning algorithm is proposed for the design of neural network-parameterized controllers. Classical model-based control methods, which include sampling- and lattice-based algorithms and model predictive control, suffer from the trade-off between model complexity and computational burden required for the online solution of expensive optimization or search problems at every short sampling time. To circumvent this trade-off, a 2-step procedure is motivated: first learning of a controller during offline training based on an arbitrarily complicated mathematical system model, before online fast feedforward evaluation of the trained controller. The contribution of this paper is the proposition of a simple gradient-free and model-based algorithm for deep reinforcement learning using task separation with hill climbing (TSHC). In particular, (i) simultaneous training on separate deterministic tasks with the purpose of encoding many motion primitives in a neural network, and (ii) the employment of maximally sparse rewards in combination with virtual velocity constraints (VVCs) in setpoint proximity are advocated.  
\end{abstract}

\section{Introduction\label{sec_intro}}

There exists a plethora of motion planning and control techniques for self-driving vehicles \cite{paden2016survey}. The diversity is caused by a core difficulty: the trade-off between model complexity and permitted online computation at short sampling times. Three popular control classes and recent vision-based end-to-end solutions are briefly summarized below.

\subsection{Model-based control methods\label{subsec_Intro_mdlBased}}

In \cite{karaman2011anytime} a \emph{sampling-based} anytime algorithm RRT$^*$ is discussed. Key notion is to refine an initial suboptimal path while it is followed. As demonstrated, this is feasible when driving towards a static goal in a static environment. However, it may be problematic in dynamic environments requiring to constantly replan paths, and where an online sampled suitable trajectory may not be returned in time. Other problems of online sampling-based methods are a limited model complexity and their tendency to produce jagged controls that require a smoothing step, e.g., via conjugate gradient \cite{dolgov2010path}. In \cite{mcnaughton2011motion}, a \emph{lattice-based} method is discussed. Such methods, and similarly also based on  \emph{motion primitives} \cite{frazzoli1999hybrid,schouwenaars2003robust,
gray2012predictive,liniger2015optimization}, are always limited by the size of the look-up table that can be searched in real-time. In \cite{mcnaughton2011motion}, a GPU is used for search. In \cite{falcone2007predictive}, \emph{linear time-varying model predictive control} (LTV-MPC) is discussed for autonomous vehicles. While appealing for its ability to incorporate constraints, MPC must trade-off model-complexity vs. computational burden from solving optimization problems online. Furthermore, MPC is dependent on state and input reference trajectories, typically for linearization of dynamics, but almost always also for providing a tracking reference. Therefore, a two-layered approach is often applied, with motion planning and tracking as the 2 layers \cite{paden2016survey}. See \cite{plessen2017spatial} for a method using geometric corridor planning in the first layer for reference generation and for the combinatorial decision taking on which side to overtake obstacles. As indicated in \cite[Sect. V-A]{falcone2007predictive} and further emphasized in \cite{plessen2017trajectory2}, the selection of reference velocities can become problematic for time-based MPC and motivated to use \emph{spatial-based} system modeling. Vehicle dynamics can be incorporated by inflating obstacles \cite{gray2012predictive}. For tight maneuvering, a linearization approach \cite{plessen2017trajectory} is more accurate, however, computationally more expensive. To summarize, 2 core observations are made. First, all methods (from sampling-based to MPC) are derived from vehicle \emph{models}. Second, all of above methods suffer from the real-time requirement of short sampling times. As a consequence, all methods make simplifications on the employed model. These include, e.g., omitting of dynamical effects, tire dynamics, vehicle dimensions, using inflated obstacles, pruning search graphs, solving optimization problems iteratively, or offline precomputing trajectories. 

\begin{figure}
\vspace{0.3cm}
\centering
\begin{tikzpicture}
\node[color=black] (a) at (-1, 0.88) { \small{human route} };
\node[color=black] (a) at (-1.23, 0.57) { \small{selection} };
\draw[->,>=latex'] (-0.05, 0.72) -- (0.75,0.72);
\draw[fill=white,solid] (0.75,0.45) rectangle (1.65,0.95);
\node[color=black] (a) at (1.2, 0.72) { Navi };
\draw[->,>=latex'] (1.2, 0.45) -- (1.2,0.15);
\draw[fill=white,solid] (0.75,-0.45) rectangle (1.65,0.15);
\node[color=black] (a) at (1.2, -0.15) { Filter };
\draw[->,>=latex'] (1.65, -0.15) -- (2.65,-0.15);
\node[color=black] (a) at (2.12, 0.03) { \small{\textbf{$s_t$}}};
\draw[fill=white,thick] (2.65,-0.45) rectangle (3.55,0.15);
\node[color=black] (a) at (3.1, -0.15) { \textbf{C} };
\node[color=black] (a) at (4, 0) { \small{\textbf{$a_t$}}};
\draw[->,>=latex'] (3.55, -0.15) -- (4.4,-0.15);
\draw[fill=white,solid] (4.4,-0.5) rectangle (5.4,0.2);
\draw[fill=white,solid] (4.6,-0.3) rectangle (5.2,-0.15);
\draw (4.74,-0.35) circle [radius=0.05];
\draw (5.04,-0.35) circle [radius=0.05];
\draw[-] (4.65, -0.15) -- (4.68,-0.02) -- (4.9,-0.02) -- (5.066,-0.15);
\draw[-] (4.85, -0.15) -- (4.85,-0.02);
\draw[->,>=latex'] (5.4, -0.15) -- (5.9,-0.15) -- (5.9,-0.75) -- (-0.05,-0.75) -- (-0.05,0.02)--(0.75,0.02);
\draw[->,>=latex'] (-0.05, -0.32) -- (0.75,-0.32);
\node[color=black] (a) at (2.95, -1.0) { \small{extero- \& proprioceptive measurements}};
\end{tikzpicture}
\caption{Closed-loop control system architecture. ``Navi'' and ``Filter'' (not focus of this paper) map human route  selections as well as extero- and proprioceptive measurements to feature vector $s_t$. This paper proposes a simple gradient-free algorithm (TSHC) for learning of controller C, which maps feature vector $s_t$ to control action $a_t$ to be applied to the vehicle.}
\label{fig:closed_loop}
\vspace{-0.6cm}
\end{figure}
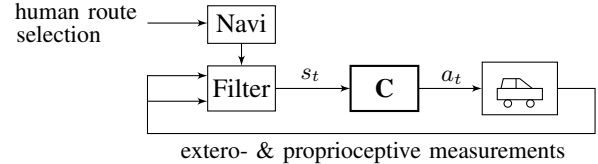

\subsection{Vision-based methods\label{subsec_IntroVision}}

In \cite{pomerleau1989alvinn} a pioneering end-to-end trained neural network labeled ALVINN was used for steering control of an autonomous vehicle. Video and range measurements are fed to a fully connected (FC)-network with a single hidden layer with 29 hidden units, and an output layer with 45 direction output units, i.e., discretized steering angles, plus one road intensity feedback unit. ALVINN does not control velocity and is trained using supervised learning based on road ``snapshots''. Similarly, recent DAVE-2 \cite{bojarski2016end} also only controls steering and is trained supervisedly. However, it outputs continuous steering action and is composed of a network including convolutional neural networks (CNN) as well as FC-layers with a total of 250000 parameters. During testing (i.e., after training), steering commands are generated from only a front-facing camera. Another end-to-end system based on only camera vision is presented in \cite{chen2017brain}. First, a driving intention (change to left lane, change to right lane, stay in lane and break) is determined, before steering angle is output from a recurrent neural network (RNN). Instead of mapping images to steering control, in \cite{chen2015deepdriving} and \cite{xu2016end}, affordance indicators (such as distance to cars in current and adjacent lanes etc.) and feasible driving actions (such as straight, stop, left-turn, right-turn) are output from neural networks, respectively. See also \cite{paxton2017combining} and their treatment of ``option policies''. To summarize, it is distinguished between (i) vision-based end-to-end control, and (ii) perception-driven approaches that attempt to extract useful features from images. Note that such features (e.g., obstacle positions) are implicitly required for all methods from Sect. \ref{subsec_Intro_mdlBased}.

\subsection{Motivation and Contribution}

This work is motivated by the following additional considerations. As noted in \cite{urmson2008autonomous}, localization relative to lane boundaries is more important than with respect to GPS-coordinates, which underlines the importance of lasers, lidars and cameras for automated driving. Second, vehicles are man- and woman-made products for which there exist decade-long experience in vehicle dynamics modeling \cite{gillespie1997vehicle},\cite{rajamani2011vehicle}. There is no reason to a priori entirely discard this knowledge (for manufacturers it is present even in form of construction plans). This motivates to leverage available vehicle models for control design. Consider also the position paper \cite{glasmachers2017limits} for general limitations of end-to-end learning. Third, a general purpose control setup is sought avoiding to switch between different vehicle models and algorithms for, e.g., highway driving and parking. There also exists only one real-world vehicle. In that perspective, a complex vehicle model encompassing all driving scenarios is in general preferable for control design. Also, a model mismatch on the planning and tracking layer can incur paths infeasible to track \cite{gray2012predictive}. Fourth, the most accident causes involving other mobile vehicles are rear-end collisions \cite{national2014traffic}, which most frequently are caused by inattentiveness or too close following distances. Control methods that enable minimal sampling times, such as feedforward control, can deterministically increase safety through minimal reaction times. In contrast, environment motion prediction (which can also increase safety) always remains stochastic. Fifth, small sampling times may contradict using complex vehicle models for control when applied for expensive online optimization or search problems. These considerations motivate a 2-step procedure: first learning of a controller during offline training based on an arbitrarily complicated mathematical system model, before online fast evaluation of the trained controller. In an automated vehicles settings, it implies that once trained, low-cost embedded hardware can be used online for evaluation of only few matrix vector multiplications.

The contribution of this paper is a simple gradient-free algorithm for model-based deep reinforcement learning using task separation with hill climbing (TSHC). Therefore, it is specifically proposed to (i) simultaneously train on separate deterministic tasks with the purpose of encoding motion primitives in a neural network, and (ii) during training to employ maximally sparse rewards in combinations with virtual velocity constraints (VVCs) in setpoint proximity. 

This paper is organized as follows. Problem formulation, the proposed training algorithm and numerical simulation experiments are discussed in Sections \ref{sec_problFormulation}-\ref{sec_expts}, before concluding.

\section{Problem Formulation and Preliminaries\label{sec_problFormulation}}

\subsection{General setup}

The problem formulation is visualized in Fig. \ref{fig:closed_loop}. Exteroceptive measurements are assumed to include inter-vehicular communication (car-2-car) sensings as well as the communication with a centralized or decentralized coordination service such that, in general, multi-automated vehicle coordination is also enabled~\cite{plessen2016multi}. For learning of controller C it is distinguished between 5 core aspects: the system model used for training, the neural network architecture used for function approximation, the training algorithm, the training tasks selection and the hardware/software implementation. Fundamental objective is to encode many desired motion primitives  (training tasks) in a neural network. The main focus of this paper is on the training algorithm aspect, motivated within the context of motion planning for autonomous vehicles characterized by nonholonomic system models.

\subsection{Illustrative system model for simulation experiments}

For simplicity a simple Euler-discretized nonlinear kinematic bicycle model~\cite{rajamani2011vehicle} is assumed for simulation experiments of Sect. \ref{sec_expts}. Equations of motion are $x_{t+1} = x_t + T_s v_t \cos(\psi_t)$, $y_{t+1}=y_t+T_sv_t \sin(\psi_t)$, $\psi_{t+1}=\psi_t + T_s \frac{v_t}{l_f}\tan(\delta_t)$, with 3 states (position-coordinates and heading), 2 controls (steering angle $\delta$ and velocity $v$), 1 system parameter (wheelbase $l_f=3.5$m), and $t$ indexing sampling time $T_s$. Coordinates $x_t$ and $y_t$ describe the center of gravity (CoG) in the inertial frame and $\psi_t$ denotes the yaw angle relative to the inertial frame. Physical actuator absolute and rate constraints are treated as part of the vehicle model on which the network training is based on. Thus, the continuous control vector is defined as $a_t = [v_t,~\delta_t]$, with $\max( v_{t-1} + \dot{v}_{\min,t}T_s,~ v_{\min,t})\leq v_t \leq \min( v_{t-1} + \dot{v}_{\max,t}T_s,~v_{\max,t})$, and $\max( \delta_{t-1} + \dot{\delta}_{\min}T_s, \delta_{\min}) \leq \delta_t \leq \min( \delta_{t-1} + \dot{\delta}_{\max}T_s,  \delta_{\max})$. The minimum velocity is negative to permit reverse driving.

\subsection{Comments on feature vector selection $s_t$}

While the mathematical system model used for training prescribes $a_t$, this is not the case for feature vector $s_t$. The dimension of $s_t$ may in general be much smaller than the system's state space. In general, $s_t$ may be an arbitrary function of filtered extero- and proprioceptive measurements according to Fig. \ref{fig:closed_loop}. Thus, a plethora of many different sensors may be compressed through the filtering to a low-dimensional $s_t$. Due to curse of dimensionality low-dimensional $s_t$ are favorable, since the easiest way to generate training tasks is to grid over the elements of $s_t$. Note further that for our purpose of encoding specific motion primitives, feature vector $s_t$ must always relate the current vehicle state with reference to a goal state (e.g., via a difference operator). Certificates about learnt control performance can be provided by statement of (i) the system model used for training, and (ii) the encoded motion primitives (training tasks) and their associated feature vectors. Ultimately, instead of only a single time-instant, $s_t$ may, in general, also represent a collection of multiple past time measurements (time-series) leading up to time $t$.

\subsection{Comments on computation\label{subsec_implementability}}

For perspective, deep learning using neural networks as function approximators is in general computationally very demanding. To underline remarkable dimensions and computational efforts in practice, note that, for example, in \cite{salimans2017evolution} training is distributed on 80 machines and 1440 CPU cores. In \cite{akiba2017extremely}, even more profoundly, 1024 Tesla P100 GPUs are used in parallel. For perspective, one Tesla P100 permits a double-precision performance of 4.7 TeraFLOPs \cite{teslaP100}.

\section{Training Algorithm\label{sec_trainalg}}

This section motivates a simple gradient-free algorithm for learning  of neural network controllers according to Fig. \ref{fig:closed_loop}.

\subsection{Neural network controller parametrization\label{subsec_netwCtrller}}


The controller in Fig. \ref{fig:closed_loop} may be parameterized by any of, e.g., FCs, LSTM cells including peephole connections \cite{gers2002learning}, GRUs \cite{cho2014learning} and variants. All neural network parameter weights to be learnt are initialized by Gaussian-distributed variables with zero mean and a small standard deviation (e.g., $0.001$). Exceptions are adding a 1 to the LSTM’s forget gate biases for LSTM cells, as recommended in \cite{jozefowicz2015empirical}, which are thus initialized with mean $1$. In proposed setting, the affine part of all FC-layers is followed by nonlinear tanh$(\cdot)$ activation functions acting elementwise. Because of their bounded outputs, saturating nonlinearities are preferred over ReLUs, which are used for the hidden layers in other RL settings \cite{lillicrap2015continuous}, but can result in large unbounded layer output changes. Before entering the neural network $s_t$ is normalized elementwise (accounting for the typical range of feature vector elements). The final FC-layer comprises a tanh$(\cdot)$ activation. It accordingly outputs bounded continous values, which are then affinely scaled to $a_t$ via physical actuator absolute and rate constraints valid at time $t$.

So far, continuous $a_t$ was assumed. A remark with respect to gear selection is made. Electric vehicles,  which appear suitable to curb urban pollution, do not require gearboxes. Nevertheless, in general $a_t$ can be extended to include \emph{discrete} gear as an additional decision variable. Suppose $N_\text{gears}$ gears are available. Then, the output layer can be extended by $N_\text{gears}$ channels, with each channel output representing a normalized probability of gear selection as a function of $s_t$, that can be trained by means of a softmax classifier.

\subsection{Reward shaping\label{subsec_rt}}

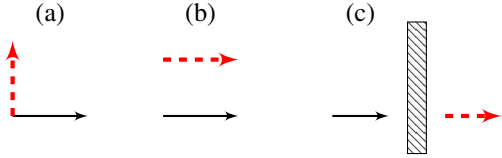
\begin{figure}
\vspace{0.3cm}
\centering
\begin{tikzpicture}
\draw[->,>=latex',thick] (0, 0) -- (1, 0);
\draw[->,>=latex',color=red,dashed, ultra thick] (0, 0) -- (0, 1);
\node[color=black] (a) at (0.5, 1.35) {(a)};
\draw[->,>=latex',thick] (2, 0) -- (3, 0);
\draw[->,>=latex',color=red,dashed, ultra thick] (2, 0.75) -- (3, 0.75);
\node[color=black] (a) at (2.5, 1.35) {(b)};
\draw[->,>=latex',thick] (4.25, 0) -- (5, 0);
\draw[->,>=latex',color=red,dashed, ultra thick] (5.75, 0) -- (6.5, 0);
\draw[pattern=north west lines, pattern color=black!70] (5.25,-0.5) rectangle (5.5,1.25);
\node[color=black] (a) at (4.625, 1.35) {(c)};
\end{tikzpicture}
\caption{The problematic of rich rewards. Three scenarios (a), (b) and (c) indicating different start (black) and goal (red dashed) states (position and heading). For (c), an obstacle is added. See Sect. \ref{subsec_rt} for discussion.}
\label{fig:ex_rewardShaping}
\vspace{-0.5cm}
\end{figure}

Reward shaping is crucial for the success of learning by reinforcement signals \cite{sutton1998reinforcement}. However, reward shaping was found to be a far from trivial matter in practical problems. Therefore, our preferred choice is motivated in detail. In most practical control problems, a state $z_0$ is given at current time $t=0$, and a desired goal state  $z^\text{goal}$ is known. Not known, however, is the shape of the best trajectory (w.r.t. a given criterion) and the control signals that realize that trajectory. Thus, by nature these problems offer a \emph{sparse} reward signal, $r_{\tilde{T}}(z^\text{goal})$, received only upon reaching the desired goal state at some time $\tilde{T}>0$. In the following, alternative \emph{rich reward} signals and \emph{curriculum learning} \cite{bengio2009curriculum} are discussed.

\subsubsection{The problematic of designing rich reward signals}

A reward signal $r_t(z_t,a_t,s_t)$, abbreviated by $r_t$, is labeled as \emph{rich} when it is time-varying as a function of states, controls and feature vector. Note that the design of any such signal is heuristic and motivated by the hope for accelerated learning through maximally frequent feedback. In the following, the problematic of rich rewards is exposed. First, let $e_{d,t} = \sqrt{ (x_t-x^\text{goal})^2 + (y_t-y^\text{goal})^2 }$, $e_{\psi,t}=|\psi_t - \psi^\text{goal} |$, and $e_{v,t} = |v_t - v^\text{goal} |$ relate states with desired goals, and let a binary flag indicate whether the desired goal pose is reached,
\begin{equation}
f_t^\text{goal} = \begin{cases} 1, & \text{if}~(e_{d,t}<\epsilon_d) \land (e_{\psi,t}<\epsilon_\psi) \land (e_{v,t}<\epsilon_v),\\
0, & \text{otherwise},
\end{cases}
\label{eq_ftgoal}
\end{equation}
where $(\epsilon_d,\epsilon_\psi,\epsilon_v)$ are small tolerance hyperparameters. Then, suppose a rich reward signal of the form $r_t = -(\alpha e_{d,t} + e_{\psi,t})$ is designed, which characterizes a weighted linear combination of different measures. This class of reward signals, trading-off various terms and providing feedback at \emph{every} $t$, occurs frequently in the literature \cite{randlov1998learning, koutnik2014online, lillicrap2015continuous, heess2017emergence}. However, as will be shown, for trajectory planning in an automotive setting (especially due to nonholonomic vehicle models), it may easily lead to undesirable behavior. Suppose case (a) in Fig. \ref{fig:ex_rewardShaping} and a maximum simulation time $T^\text{max}$. Then, omitting a discount factor for brevity, $-T^\text{max}\frac{\pi}{2} > - \sum_{t=0}^{\tilde{T}}\alpha e_{d,t}^\star +  e_{\psi,t}^\star$, may be obtained for accumulated rewards. Thus, the no-movement solution may incur more accumulated reward, namely $-T^\text{max}\frac{\pi}{2}$, in comparison to the true solution, which is indicated on the right-hand side of the inequality sign. 

Similarly, for specific $(\alpha,T^\text{max})$, the second scenario (b) in Fig. \ref{fig:ex_rewardShaping} can return a no-movement solution since the initial angle is already coinciding with the target angle. Hence, for a specific $(\alpha,T^\text{max})$-combination, the accumulated reward when not moving may exceed the value of the actual solution.

The third scenario (c) in Fig. \ref{fig:ex_rewardShaping} shows that even if reducing rich rewards to a single measure, e.g., $r_t=-e_{d,t}$, an undesired standstill may result. This occurs especially in the presence of obstacles (and maze-like situations in general).

To summarize, for finite $T^\text{max}$, the design of rich reward signals is not straightforward and can easily result in solution trajectories that may even be globally optimal w.r.t. accumulated reward, however, prohibit to solve the original problem of determining a trajectory from initial to target state.     

\subsubsection{The problematic of curriculum learning}

In \cite{bengio2009curriculum}, curriculum learning (CL) is discussed as a method to speed up learning by providing the learning agent first with simpler examples before gradually increasing complexity. Analogies to humans and animals are drawn. The same paper also acknowledges the difficulty of determining ``interesting'' examples \cite[Sect. 7]{bengio2009curriculum} that optimize learning progress.

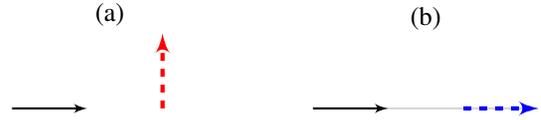
\begin{figure}
\vspace{0.3cm}
\centering
\begin{tikzpicture}
\draw[->,>=latex',thick] (0, 0) -- (1, 0);
\draw[->,>=latex',color=red,dashed, ultra thick] (2, 0) -- (2, 1);
\node[color=black] (a) at (1.3, 1.25) {(a)};
\draw[->,>=latex',color=black!30] (4, 0) -- (7, 0);
\draw[->,>=latex',thick] (4, 0) -- (5, 0);
\draw[->,>=latex',color=blue,dashed, ultra thick] (6, 0) -- (7, 0);
\node[color=black] (a) at (5.5, 1.2) {(b)};
\end{tikzpicture}
\caption{The problematic of curriculum learning. The difficulty of selecting ``simple'' examples is illustrated, see Sect. \ref{subsec_rt} for discussion. The original problem with start (black) and goal (red dashed) state is denoted in (a). A ``simpler'' problem is given in (b).}
\label{fig:ex_curriculumLearning}
\vspace{-0.5cm}
\end{figure}

Indeed, CL entails the following issues. First, ``simpler'' tasks need to be identified. This is not straightforward as discussed shortly. Second, these tasks must first be solved before their result can serve as initialization to more complex tasks. In contrast, without CL, the entire solution time can be devoted to the complex tasks rather than being partitioned into easier and difficult tasks. In experiments, this was found to be relevant. Third, the solution of an easier task does not necessarily represent a better initialization to a harder problem in comparison to an alternative random initialization. For example, consider the scenario in Fig. \ref{fig:ex_curriculumLearning}. The solution of the simpler task does not serve as a better initialization than a purely random initialization of weights. This is since the final solution requires outreaching steering and possibly reversing of the vehicle. The simpler task just requires forward driving and stopping. This simple example illustrates the need for careful manual selection of suitable easier tasks for CL.

\subsubsection{The benefits of maximal sparse rewards in combination with virtual velocity constraints}

In the course of this work, many reward shaping methods were tested. These include, first, solving ``simpler'' tasks by first dismissing target angles limited to $30^\circ$-deviation from the initial heading. Second, $\epsilon$-tolerances were initially relaxed before gradually decreasing them. Third, it was tested to first solve a task for only the $\epsilon_d$-criterion, then both $(\epsilon_d,\epsilon_\psi)$, and only finally all of $(\epsilon_d,\epsilon_\psi, \epsilon_v)$. Here, also varying sequences (e.g., first $\epsilon_\psi$ instead of $\epsilon_d$) were tested. No consistent improvement could be observed for neither of these methods. On the contrary, solving allegedly simpler task reduced available solver time for the original ``hard'' problems. Without CL the entire solution time can be devoted to the complex tasks. 

Based on these findings, our preferred reward design method is maximally sparse and defined by
\begin{align}
r_t &= \begin{cases} -\infty, & \text{if}~f_t^\text{crash}==1,\\
-1, & \text{otherwise},   \end{cases}\label{eq_def_rt}\\
F_i &= \begin{cases} 1, & \text{if}~\sum_{\tau=t-T^\text{goal}+1}^t f_{\tau}^\text{goal}==T^\text{goal},\\
0, & \text{otherwise},\label{eq_def_Fi}
\end{cases}
\end{align}
where $f_t^\text{goal}$ from \eqref{eq_ftgoal}, and  $f_t^\text{crash}$ being an indicator flag for a vehicle crash. Thus, upon $F_i=1$ the RL problem is considered as solved. In addition, the pathlength incurred for a transition from sampling time $t$ to $t+1$ is defined as
\begin{equation}
\Delta p_t = -\sqrt{ (x_{t+1}-x_t)^2 + (y_{t+1}- y_t)^2 }.
\end{equation}
As elaborated below, accumulated total pathlength is used to rank solution candidates solving all desired training tasks.   

The integral for $F_i$ is defined for generality, in particular for problems such as the inverted pendulum \cite{anderson1989learning} in mind, which are considered to be solved only after stabilization is demonstrated for sufficiently many consecutive $T^\text{goal}$ time steps. Note, however, that this is not required for an automotive setting. Here, it must be $T^\text{goal}=1$. Only then learning with $v^\text{goal}\neq 0$ is possible. Other criteria and trade-offs for $\Delta p_t$ are possible (e.g., accumulated curvature of resulting paths and a minmax objective therefore). The negation is introduced for maximization (``hill climbing''-convention). Note that the preferred reward signal is maximal sparse, returning $-1$, for all times up until reaching the target. It represents a \emph{tabula rasa} solution critizised in \cite{randlov1998learning} for its maximal sparsity. Indeed, standalone it was not sufficient to facilitate learning when also accounting for a velocity target $v^\text{goal}$. Therefore, \emph{virtual velocity constraints} (VVCs) in target proximity are introduced. Two variants are discussed. First, VVCs spatially dependent on $e_{d,t}$ can be defined as
\begin{equation}
v_t^m = \begin{cases} v^m, & \text{if}~e_{d,t}\geq R_{v}^\text{thresh},\\
v^\text{goal} + \frac{(v^m - v^\text{goal})}{R_{v}^\text{thresh}} e_{d,t}, & \text{otherwise},
\end{cases}
\label{eq_virtualCstrt_v}
\end{equation}
where $m\in\{\text{max},\text{min}\}$, and $R_{v}^\text{thresh}$ is a hyperparameter (e.g., range-view length or a heuristic constant). Second and alternatively, VVCs may be defined as spatially invariant with a constant margin (e.g., 5km/h) around the target velocity. For both variants, the neural network output that regulates velocity is scaled with updated $v_t^\text{min}$ and $v_t^\text{max}$ constraints (i.e., using  \eqref{eq_virtualCstrt_v} for spatially dependent VVCs).  

Let us further legitimize VVCs. Since speed is a decision variable it can always be constrained artificially. This justifies the introduction of VVCs. In \eqref{eq_virtualCstrt_v}, bounds are set to affinely converge towards $v^\text{goal}$ in the proximity of the goal location. This is a heuristic choice. Note that the affine choice do \emph{not} necessarily imply constant accelerations. This is since \eqref{eq_virtualCstrt_v} is spatially parameterized. Note further that physical actuator rate constraints still hold when $a_t$ is applied to the vehicle. 

It was also tested to constrain $\delta_t$.  The final heading pose implies circles prohited from trespassing because of the nonholonomic vehicle dynamics. It was tested to add these as virtual obstacles. However, this did not accelerate learning.

Finally, note that VVCs artificially introduce hard constraints and thus shape the learning result w.r.t velocity, at least towards the end of the trajectory. Two comments are made. First, in receding online operation, with additional frequent resetting of targets, this shaping effect is reduced since only the first control of a planned trajectory is applied. Second, in case of spatially dependent VVCs the influence of hyperparameter $R_v^\text{thres}$ only becomes apparent during parking when following the trajectory up until standstill. Here, however, no significant velocity changes are desired, such that the $R_v^\text{thres}$-choice is not decisive. Ultimately, note that sparse rewards naturally avoid the need to introduce trade-off hyperparameters for the weighting of states in different units. This permits solution trajectories between start and goal poses to naturally evolve without biasing them by provision of rich references to track.

To summarize this section. It was illustrated that the design of \emph{rich reward} signals as well as \emph{curriculum learning} can be problematic. Therefore, \emph{maximal sparse rewards} in combination with \emph{virtual velocity constraints} are proposed. 

\subsection{The role of tolerances $\epsilon$\label{subsec_role_tolerances_eps}}

Tolerances $\epsilon$ in \eqref{eq_ftgoal} hold an important role for 2 reasons. On one hand, nonzero $\epsilon$ result in deviations between actually learnt $\hat{z}^\text{goal}$ and originally desired goal pose $z^\text{goal}$. On the other hand, very small $\epsilon$ (e.g., $\epsilon_d=0.1$m, $\epsilon_\psi=1^\circ$ and $\epsilon_v=1$km/h) prolong learning time. Two scenarios apply.

First, for a network trained on a large-scale and dense grid of training tasks and for small $\epsilon$, during online operation, suitable control commands are naturally interpolated even for setpoints not seen during training. The concept of natural interpolation through motion primitives encoded in neural networks is the core advantage over methods relying on look-up tables with stored trajectories, which require to solve time-critical search problems. For example, in \cite{mcnaughton2011motion} exhaustive search of the entire lattice-graph is conducted online on a GPU. In \cite{liniger2015optimization}, a total of about 100 motion primitives is considered. Then, online an integer program is solved by enumeration using maximal progress along the centerline as criterion for selection of the best motion primitive. In contrast, for control using neural networks as function approximators this search is not required. 

Second, the scenario was considered in which existing training hardware does (i) not permit large-scale encoding, and (ii) only permits to use larger $\epsilon$-tolerances to limit training time. Therefore, the following method is devised. First, tuples $(\hat{z}^\text{goal},z^\text{goal})$ are stored for each training task. Then, during online operation, for any setpoint, $z^\text{setpoint}$, the closest (according to a criterion) $\hat{z}^\text{goal}$ from the set of training tasks is searched, before the corresponding $z^\text{goal}$ is applied to the network controller. Two comments are made. First, in order to reach $\hat{z}^\text{goal}$ (with zero deviation), $z^\text{goal}$ must be applied to the network. Therefore, \emph{tuples} need to be stored. Second, eventhough this method now also includes a search, it still holds an important advantage over lattice-based methods. This is the compression of the look-up table in the network weights. Hence, only tuples need to be stored---not entire trajectories. This is especially relevant in view of limited hardware memory. Thus, through encoding, potetially many more motion primitives can be stored. 

In practice, the first scenario is preferable. It is also implementable for 2 reasons. First, see Sect. \ref{subsec_implementability} for computational opportunities. Second, neural networks have in principle unlimited function approximation capability~\cite{siegelmann1991turing}. Hence, the implementation of the first approach is purely a question of intelligent task setup, and computational power. 

\subsection{Main Algorithm -- TSHC\label{subsec_mainAlg_TSHC}}

Algorithm \ref{alg_TSHC} is proposed for simple gradient-free model-based reinforcement learning. The name is derived from the fact of (i) learning from separate training tasks, and (ii) a hill climbing update of parameters (greedy local search).

Let us elaborate on definitions. Analysis is provided in Sect. \ref{subsec_analysis_TSHC}. First, all network parameters are lumped into variable $\theta$. Second, the perturbation step 8 in Algorithm \ref{alg_TSHC} has to be intepreted accordingly. It implies parameter-wise affine perturbations with zero-mean Gaussian noise and  spherical variance $\sigma_\text{pert}^2$. Third, $\mathcal{X}(\cdot)$, $\mathcal{R}(\cdot)$ and $\mathcal{Z}(\cdot)$ in Steps 14-16 denote functional mappings between properties defined in the preceding sections. Fourth, hyperparameters are stated in Step 1. While $N_\text{restarts}$, $N_\text{iter}^\text{max}$, $n$, $N_\text{tasks}$ and $T^\text{max}$ denote lengths of different iterations, $\beta>1$ is used for updating of $\sigma_\text{pert}$ in Step 35 and 37. Fifth, for every restart iteration, $i_\text{restart}$, multiple parameter iterations are conducted, at most $N_\text{iter}^\text{max}$ many. Sixth, in Steps 25 and 29 hill climbing is conducted, when (i) all tasks have been solved for current $i_\text{iter}$, or (ii) not all tasks have yet been solved, respectively. Seventh, there are 2 steps in which an early termination of iterations may occur: Step 21 and 41. The former is a must. Only then learning with $v^\text{goal}\neq 0$ is possible. The latter termination criterion in Step 41 is optional. If dismissed, a refinement step is implied. Thus, eventhough all $N_\text{tasks}$ tasks have been solved, parameter iterations (up until $N_\text{iter}^\text{max}$) are continued. Eighth, note that a discount hyperparameter $\gamma$, common to gradient-based RL methods \cite{schulman2017proximal}, is not required. This is since it is irrelevant in the maximally sparse reward setting. Ninth, nested parallelization is in principle possible with an inner and outer parallelization of Steps 10-22 and 7-22, respectively. The former refers to $N_\text{tasks}$ solutions for a given parameter vector $\theta_i$, whereas the latter parallelizes $n$ parameter perturbations. For final experiments, Steps 7-22 were implemented asynchronously. Finally, there are 3 options considered for $\sigma_\text{pert}$-selection. First, holding an initial $\sigma_\text{pert}$-selection constant throughout TSHC. Second, updating $\sigma_\text{pert}$ randomly (e.g., uniformly distributed between 10 and 1000), whereby this can be implemented either in Step 4 at every $i_ \text{restart}$, or in Step 6 at every $(i_\text{restart},i_\text{iter})$-combination. Third, $\sigma_\text{pert}$ may be adapted according to progress in $N^{\text{tasks},\star}$, as outlined in Algorithm \ref{alg_TSHC}. For the first 2 options of selecting $\sigma_\text{pert}$, Steps 34-37 are dismissed and at least $\beta$ can be dismissed from the list of hyperparameters in Step 1. 

\vspace{-0.2cm}
\begin{algorithm}
\begin{small}
\DontPrintSemicolon
\textbf{Input:} system model, network structure, $N_\text{tasks}$ training tasks; $N_\text{restarts}$, $N_\text{iter}^\text{max}$, $n$, $T^\text{max}$, $\epsilon$; and a method to update $\sigma_\text{pert}$: constant, random or adaptive based on ($\beta$,  $\sigma_\text{pert}^\text{min}$, $\sigma_\text{pert}^\text{max}$).\;
Initialize $\theta^\star \leftarrow \emptyset$, $N^\star\leftarrow 0$, $P^\star \leftarrow -\infty$, $J^\star \leftarrow 0$.\;
\For{$i_\text{restart}=1,\dots,N_\text{restarts}$}
{
Initialize $\theta$ randomly, and $\sigma_\text{pert}\leftarrow \sigma_\text{pert}^\text{max},~N_\text{old}^{\text{tasks},\star}\leftarrow 0$.\;
\For{$i_\text{iter}=1,\dots,N_\text{iter}^{\text{max}}$}
{
\% RUN ASYNCHRONOUSLY: \;
\For {$i=1,\dots,n$}
{
Perturb $\theta_i \leftarrow \theta + \sigma_\text{pert}\zeta$, with $\zeta\sim\mathcal{N}(0,I)$.\;
Initialize $N_i^{\text{tasks},\star}\leftarrow 0$, $P_i\leftarrow 0$, $J_i \leftarrow 0$.\;
\For {$i_\text{task}=1,\dots,N_\text{tasks}$}
{
Initialize $z_0$ (and LSTM and GRU cells).\;
\For {$t=0,\dots,T^\text{max}-1$}
{
Read $s_t$ from $i_\text{task}$-environment.\;
$a_t \leftarrow \mathcal{X}(s_t,\theta_i)$.\;
$(r_t,\Delta p_t,f_t^\text{goal}) \leftarrow \mathcal{R}(s_t,a_t)$.\;
$z_{t+1} \leftarrow \mathcal{Z}(z_t,a_t)$.\;
$P_i \leftarrow P_i + \Delta p_t$.\;
$J_i \leftarrow J_i + r_t$.\;
$F_i$ according to \eqref{eq_def_Fi}.\;
\If {$(F_i==1) \lor(r_t==-\infty)$}
{Break $t$-loop.\;}
} 
$N_i^{\text{tasks},\star} \leftarrow N_i^{\text{tasks},\star} + F_i$.\;
} 
} 
%
%
\% DETERMINE $i^\star$: \;
\If {$\max\limits_i\{N_i^{\text{tasks},\star}\}_{i=1}^n==N_\text{tasks}$}
{
$i^\star= \text{arg}\max\limits_i \left\{P_i ~|~  N_i^{\text{tasks},\star}==N_\text{tasks}\right\}_{i=1}^n$.\;
\If {$P_{i^\star}>P^\star$}{
$ (\theta^\star,N^\star,P^\star,J^\star ) \leftarrow ( \theta_{i^\star}, N_\text{tasks}, P_{i^\star},J_{i^\star})$.\;
}
}
\Else
{
$i^\star= \text{arg}\max\limits_i \left\{ J_i \right\}_{i=1}^n$.\;
\If {$(J_{i^\star}>J^\star) \land (P^\star==-\infty)$}{
$ (\theta^\star,N^\star,P^\star,J^\star ) \leftarrow ( \theta_{i^\star}, N_{i^\star}^\text{tasks}, P_{i^\star},J_{i^\star})$.\;
}
}
$N^{\text{tasks},\star} \leftarrow N_{i^\star}^{\text{tasks},\star}$.\;
%
%
\% UPDATE PARAMETERS: \;
\If {$N^{\text{tasks},\star} > N_\text{old}^{\text{tasks},\star}$}
{ 
$\sigma_\text{pert} \leftarrow \max(\frac{1}{\beta}\sigma_\text{pert},~\sigma_\text{pert}^\text{min})$.\;
}
\ElseIf {$N^{\text{tasks},\star} < N_\text{old}^{\text{tasks},\star}$}
{
$\sigma_\text{pert} \leftarrow \min(\beta\sigma_\text{pert},~\sigma_\text{pert}^\text{max})$.\;
}
$\theta \leftarrow \theta_{i^\star}$ and $N_\text{old}^{\text{tasks},\star}\leftarrow N^{\text{tasks},\star}$.\;
%
%
\% OPTIONAL: \;
\If {$N_i^{\text{tasks},\star}==N_\text{tasks}$}
{ Break $i_\text{iter}$-loop. {\color{black} \% no further refinement step.}\;}
%
%
} 
} 
\textbf{Output}: $(\theta^\star,N^\star,P^\star,J^\star )$.\;
\caption{TSHC}\label{alg_TSHC}
\end{small}
\end{algorithm}
\vspace{-0.5cm}

\subsection{Analysis \label{subsec_analysis_TSHC}}

According to classifications in \cite{fu2005simulation}, TSHC is a gradient-free instance-based simulation optimization method, generating new candiate solutions based on only the current solution and random search in its neighborhood. Because of its hill climbing (greedy) characteristic, it differs from (i) \emph{evolutionary} (population-based) methods that construct solution by combining others typically using weighted averaging \cite{wierstra2014natural, salimans2017evolution}, and (ii) from model-based methods that use \emph{probability distributions} on the space of solution candidates, see \cite{fu2005simulation} for a survey. In its high-level structure, Algorithm \ref{alg_TSHC} can be related to the COMPASS algorithm \cite{xu2010industrial}. Within a global stage, they identify several possible regions with locally optimal solutions. Then, they find local optimal solutions for each of the identified regions, before they select the best solution among all identified locally optimal solutions. In our setting, these regions are enforced as the separate training tasks and the best solution for all of  these is selected.

In combination with sufficiently large $n$, $\sigma_\text{pert}$ must be large enough to permit sufficient exploration such that a network parametrization solving all tasks can be found. In contrast, the effect of decreasing $\sigma_\text{pert}$ with an increasing number of solved tasks is that, ideally, a speedup in learning progress results from the assignment of more of solution candidates $\theta_i$ closer in variance to a promising $\theta$ (see Step 8 of TSHC). 

Steps 29-31 are discussed. For the case that for a specific $i_\text{iter}$-iteration not all tasks have yet been solved, $i^\star= \text{arg}\max\limits_i \{ N_i^{\text{tasks},\star} \}_{i=1}^n$ has been considered as an alternative criterion for Step 29. Several remarks can be made. First, Step 29 and the alternative are not equal. This is because, in general, different tasks are solved in a different number of time steps. However, the criteria are approximately equivalent for sparse rewards (since $J_i$ accumulates constants according to \eqref{eq_def_rt}), and especially for large $T^\text{max}$. The core advantage of employing Step 29 in TSHC is that it can, if desired, also be used in combination with rich rewards to accelerate learnig progress (\emph{if} a suitable rich reward signal can be generated). In such a scenario, $i^\star$ according to Step 29 is updated towards most promisining $J_{i^\star}$, then representing the accumulated rich reward. Thus, in contrast to \eqref{eq_def_rt}, a rich reward could be represented by a weighted sum of squared errors between state $z_t\in\mathbb{R}^{n_z}$ and a reference $z_t^\text{ref}\in\mathbb{R}^{n_z}$, 
\begin{equation}
r_t = \begin{cases} -\infty, & \text{if}~f_t^\text{crash}==1,\\
- \sum_{l=0}^{n_z-1} \alpha_l (z_t(l) - z_t^\text{ref}(l))^2, & \text{otherwise}, \end{cases}
\label{eq_rt_rich}
\end{equation}
where $\alpha_l$, are trade-off hyperparameters and scalar elements of vectors are indexed by $l$ in brackets. Another advantage of the design in Algorithm \ref{alg_TSHC} according to Step 29-31 is its \emph{anytime solution} character. Even if not all $N_\text{tasks}$ are solved, the solution returned for the tasks that are solved, typically is of good quality and optimized according to Steps 29-31. 

If for all $N_\text{tasks}$ tasks there exists a feasible solution for a given system model and a sufficiently expressive network structure parameterized by $\theta$, then Algorithm \ref{alg_TSHC} can find such parametrization for sufficiently large hyperparameters $N_\text{restarts}$, $N_\text{iter}^\text{max}$, $n$, $T^\text{max}$ and $\sigma_\text{pert}^\text{max}$. The solution parametrization $\theta^\star$ is the result from the initialization Step 4 and parameter perturbations according to Step 8, both nested within multiple iterations. As noted in \cite{hong2009brief}, for optimization via simulation, a global convergence guarantee provides little practical meaning other than reassuring a solution will be found ``eventually'' when simulation effort goes to infinity. However, the same reference also states that a convergence property is most meaningful if it can help in designing suitable stopping criteria. In our case, there are 2 such conceptual levels of stopping criteria: first, the solution of all training tasks, and second, the refinement of solutions. 

Control design is implemented hierarchically in 2 steps. First, suitable training tasks (desired motion primitives) are defined. Then, these are encoded in the network by the application of TSHC. This has practical implications. First, it encourages to train on deterministic tasks. Furthermore, at every $i_\text{iter}$, it is simultaneously trained on all of these separate tasks. This is beneficial in that the best parametrization, $\theta^\star$, is clearly defined via Step 25, maximizing the accumulated $P$-measure  over \emph{all} tasks. Second, it enables to provide certificates on the learnt performance, which can be provided by stating (i) the employed vehicle model, and (ii) the list of encoded tasks (motion primitives). Note that such certificates cannot be given for the class of \emph{stochastic} continuous action RL algorithms that are derived from the Stochastic Policy Gradient Theorem \cite{sutton2000policy}. This class includes all stochastic actor-critic algorithms, including A3C \cite{mnih2016asynchronous} and PPO \cite{schulman2017proximal}.

\subsection{Discussion and comparison with related RL work\label{subsec_relatedWork_TSHC}}

Related continuous control methods that use neural network for function approximation are discussed, focusing on one stochastic \cite{schulman2017proximal}, one deterministic policy gradient method \cite{lillicrap2015continuous}, and one evolution strategy \cite{salimans2017evolution}. The methods are discussed in detail to underline aspects of TSHC. 

First, the \emph{stochastic policy gradient} method PPO \cite{schulman2017proximal} is discussed. Suppose that a stochastic continus control vector is sampled from a Gaussian distribution parameterized\footnote{In this setting, mean and variance of the Gaussian distribution are the output of a neural network whose parameters are summarized by lumped $\theta$.} by $\theta$ such that $a_t\sim \pi(a_t | s_t,\theta)$. Then,
\begin{equation}
J(\theta) = \underset{s_t,a_t}{\text{E}} \left[ g(s_t,a_t\sim \pi(a_t | s_t,\theta)) \right],\label{eq_Jtheta_stochastic}
\end{equation}
is defined as the expected accumulated and time-discounted reward when at $s_t$ drawing $a_t$, and following the stochastic policy for all subsequet times when acting in the simulation environment. Since function  $g(s_t,a_t)$ is a priori not know, it is parameterized by $\theta_{V,\text{old}}$ and estimated. Using RL-terminology, in the PPO-setting, $g(s_t,a_t)$ represents the \emph{advantage function}. Then, using the ``log-likelihood trick'', and subsequently a first-order Taylor approximation of $\log( \pi(a_t | s_t,\theta) )$ around some reference $\pi(a_t | s_t,\theta_\text{old})$, the following parameterized cost function is obtained as an \emph{approximation} of \eqref{eq_Jtheta_stochastic}, 
\begin{equation}
\tilde{J}(\theta) = \underset{s_t,a_t}{\text{E}} \left[ \hat{g}_t(\theta_{V,\text{old}}) \frac{\pi(a_t | s_t,\theta)}{\pi(a_t | s_t,\theta_\text{old})} \right].
\label{eq_Jtheta_stochasticPPO_approx}
\end{equation}
Finally, \eqref{eq_Jtheta_stochasticPPO_approx} is modified to the final PPO-cost function~\cite{schulman2017proximal}
\begin{align}
\hat{J}(\theta) =& \underset{s_t,a_t}{\text{E}} \big[\min\bigg(  \hat{g}_t(\theta_{V,\text{old}}) \frac{\pi(a_t | s_t,\theta)}{\pi(a_t | s_t,\theta_\text{old})},\notag\\
& \text{clip}( \frac{\pi(a_t | s_t,\theta)}{\pi(a_t | s_t, \theta_\text{old})}, 1-\epsilon,1+\epsilon)\hat{g}_t(\theta_{V,\text{old}}) \bigg) \big],
\label{eq_Jtheta_stochasticPPO_PPO}
\end{align}
whereby the advantage function is estimated by the policy parameterized by $(\theta_{V,\text{old}},\theta_{\text{old}})$, which is run for $T$ consecutive time steps such that for all $t$ the tuples $(s_t,a_t,r_t,s_{t+1},\hat{g}_t(\theta_{V,\text{old}}))$ can be added to a \emph{replay buffer}, from which later minibatches are drawn. According to  \cite{schulman2017proximal}, the estimate is $\hat{g}_{T-1}(\theta_{V,\text{old}}) = \kappa_{T-1}$ with $\kappa_{T-1} = r_{T-1} + \gamma V(s_T,\theta_{V,\text{old}}) - V(s_{T-1},\theta_{V,\text{old}})$, $\hat{g}_{T-2}(\theta_{V,\text{old}}) = \kappa_{T-2} + \gamma \lambda \hat{g}_{T-1}(\theta_{V,\text{old}})$ and so forth until $\hat{g}_{0}(\theta_{V,\text{old}})$, and where $V(s,\theta_{V,\text{old}})$ represents a \emph{second}, the so-called critic neural network. Then, using uniform randomly drawn minibatches of size $M$, parameters $(\theta_V,\theta)$ of both networks are updated according to $\text{arg}\min\limits_{\theta_V}\frac{1}{M} \sum_{i=0}^{M-1} \left(  V(s_i,\theta_V) - \left(\hat{g}_{i}(\theta_{V,\text{old}}) - V(s_i,\theta_{V,\text{old}})\right) \right)^2$ and $\text{arg}\max\limits_{\theta}\frac{1}{M} \sum_{i=0}^{M-1} \mathcal{G}_i(\theta)$, with $\mathcal{G}_i(\theta)$ denoting the argument of the expectation in \eqref{eq_Jtheta_stochasticPPO_PPO} evaluated at time-index $i$. This relatively detailed discussion is given to underline following observations. With first the introduction of a parameterized estimator, then a first-order Taylor approximation, and then clipping, \eqref{eq_Jtheta_stochasticPPO_PPO} is an arguably crude approximation of the original problem \eqref{eq_Jtheta_stochastic}. Second, the complexity with \emph{two} actor and critic networks is noted. Typically, both are of the same dimensions apart from the output layers. Hence, when not sharing weights between the networks, approximately \emph{twice} as many parameters are required. However, \emph{when} sharing any weights between actor and critic network, then optimization function \eqref{eq_Jtheta_stochasticPPO_PPO} must be extended accordingly, which introduces another approximation step. Third, note that gradients of both networks must be computed for backpropagation. Fourth, the dependence on \emph{rich} reward signals is stressed. As long as the current policy does not find a solution candidate, in a \emph{sparse} reward setting, all $r_i$ are uniform. Hence, there is no information permitting to find a suitable parameter update direction and all of the computational expensive gradient computations are essentially not usable\footnote{It is mentioned that typically the first, for example, 50000 samples are collected \emph{without} parameter update. However, even then that threshold must be selected, and the fundamental problem still perseveres.}. Thus, the network parameters are still updated entirely at random. Moreover, \emph{even if} a solution candidate trajectory was found, it is easily averaged out through the random minibatch update. This underlines the problematic of \emph{sparse} rewards for PPO. Fifth, A3C \cite{mnih2016asynchronous} and PPO \cite{schulman2017proximal} are by nature  \emph{stochastic} policies, which \emph{draw} their controls from a Gaussian distribution (for which mean and variance are the output of a trained network with current state as its input). Hence, exact repetition of any task (e.g., the navigation between 2 locations) cannot be guaranteed. It can only be guaranteed if dismissing the variance component, and consequently using solely the mean for deterministic control. This can be done in practice, however, introduces another approximation step.

\emph{Deterministic policy gradient} method DDPG \cite{lillicrap2015continuous} is discussed. Suppose a deterministic continuous control vector parameterized such that $a_t = \mu(s_t,\theta)$.  Then, the following cost function is defined,
\begin{equation*}
J(\theta) = \underset{s_t,a_t}{\text{E}} \left[ g(s_t,a_t= \mu(s_t,\theta)) \right] = \underset{s_t}{\text{E}} \left[ g(s_t,\mu(s_t,\theta)) \right].
\end{equation*}  
Its gradient can now be computed by applying the chain-rule for derivatives \cite{silver2014deterministic}. Introducing a parameterized estimate of $g(s_t,a_t)$, which here represents the \emph{Q-function} or \emph{action value function} (in contrast to the advantage function in above stochastic setting), the final DDPG-cost function \cite{lillicrap2015continuous} is
\begin{equation*}
J(\theta) =  \underset{s_t}{\text{E}} \left[ \hat{g}(s_t,\mu(s_t,\theta),\theta_Q) \right].
\end{equation*}
Then, using minibatches, critic and actor network parameters $(\theta_Q,\theta)$ are updated as $\text{arg}\min\limits_{\theta_Q}\frac{1}{M} \sum_{i=0}^{M-1} (  \hat{g}(s_i,a_i,\theta_Q) -\left(r_i + \gamma Q(s_{i+1},\mu(s_{i+1},\theta_\text{old}),\theta_{Q,\text{old}})\right) )^2$ and  $\text{arg}\min\limits_{\theta}\frac{1}{M} \sum_{i=0}^{M-1} Q(s_i,\mu(s_i,\theta),\theta_Q)$, with slowly tracking target network parameters $(\theta_{Q,\text{old}},\theta_\text{old})$. Several remarks can be made. First, the Q-function is updated towards only its \emph{one-step} ahead target. It is obvious that rewards are therefore propagated \emph{very} slowly. For sparse rewards this is even more problematic than for rich rewards, especially because of the additional danger of averaging out important update directions though random minibatch sampling. Furthermore, and analogous to the stochastic setting, for the sparse reward setting, as long as no solution trajectory was found, all of the gradient computations are not usable and all network parameters are still updated entirely at random. DDPG is an \emph{off-policy} algorithm. In \cite{lillicrap2015continuous}, exploration of the simulation environment is achieved according to the current policy plus additive noise following an \emph{Ornstein-Uhlenbeck} process. This is a mean-reverting linear stochastic differential equation \cite{geering2011stochastic}. A first-order Euler approximation thereof can be expressed as the action exploration rule $a_t = \mu(s_t,\theta)(1-P_\theta^\text{OU}) + P_\sigma^\text{OU} \epsilon,~\epsilon\sim\mathcal{N}(0,I)$, with hyperparameters $(P_\theta^\text{OU},P_\sigma^\text{OU})=(0.15,0.2)$ in \cite{lillicrap2015continuous}. This detail is provided to stress a key difference between policy gradient methods (both stochastic and deterministic), and methods such as \cite{salimans2017evolution} and TSHC. Namely, while the former methods sample controls from the stochastic policy or according to \emph{heuristic} exploration noise before updating parameters using minibatches of \emph{incremental} tuples $(s_i,a_i,r_i,s_{i+1})$ plus $\hat{g}_i(\theta_{V,\text{old}})$ for PPO, the latter directly work in the \emph{parameter space} via local perturbations, see Step 8 of Algorithm \ref{alg_TSHC}. This approach appears particularly suitable when dealing with sparse rewards. As outlined above, in such setting, parameter updates according to policy gradient methods are also entirely at random, however, with the computationally significant difference of first an approximately four times as large parameter space and, second, the unnecessary costly solution of non-convex optimization problems as long as no solution trajectory has been found. A well-known issue in training neural networks is the problem of vanishing or exploding gradients. It is particularly relevant for networks with saturating nonlinearities and can be addressed by batch \cite{ioffe2015batch} and layer normalization \cite{ba2016layer}. In both normalization approaches, \emph{additional} parameters are introduced to the network which must be learnt (bias and gains). These issues are not relevant for the proposed gradient-free approach. 

\begin{figure*}
\centering%
\includegraphics[scale=0.06]{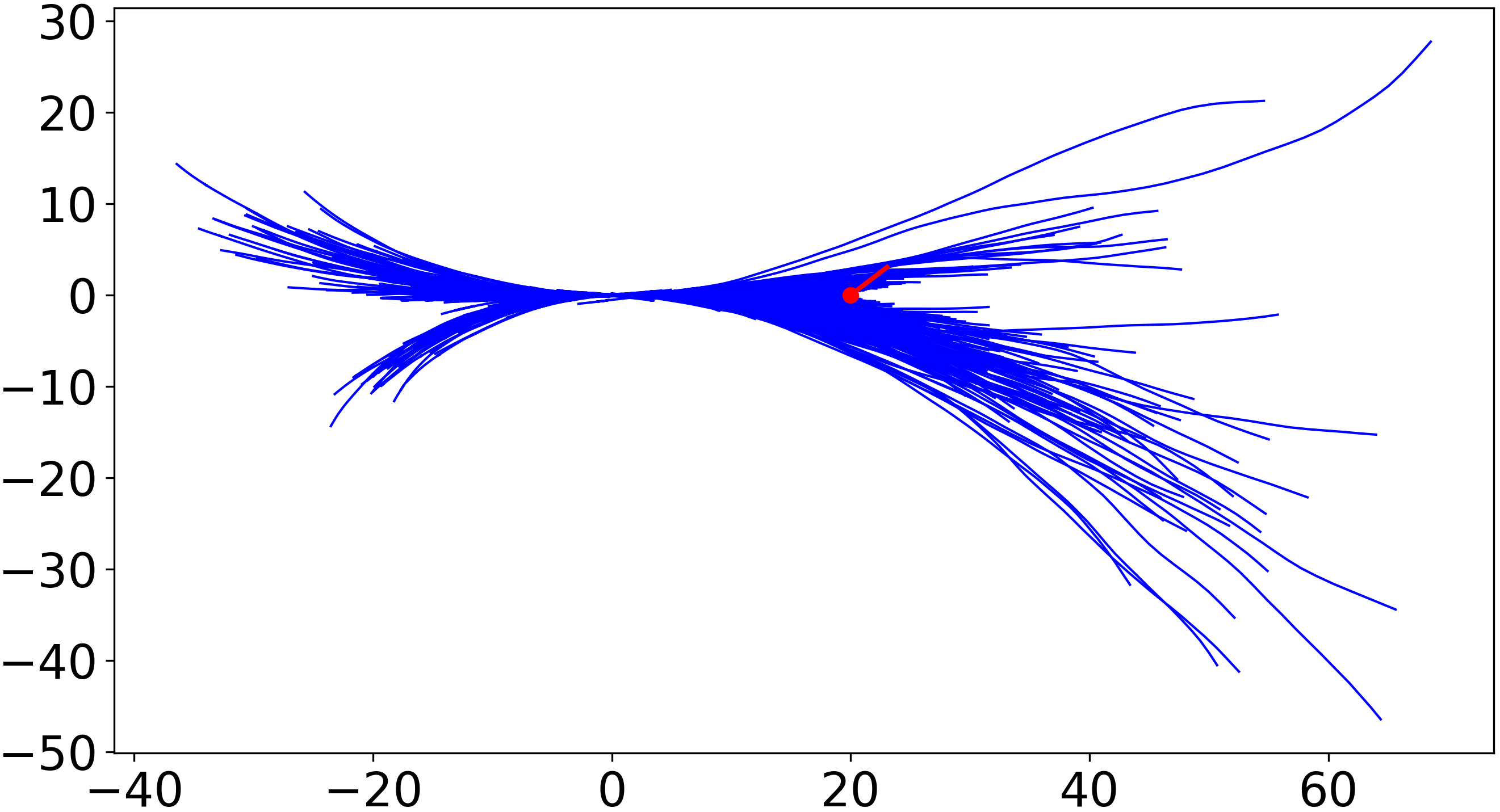}
\includegraphics[scale=0.06]{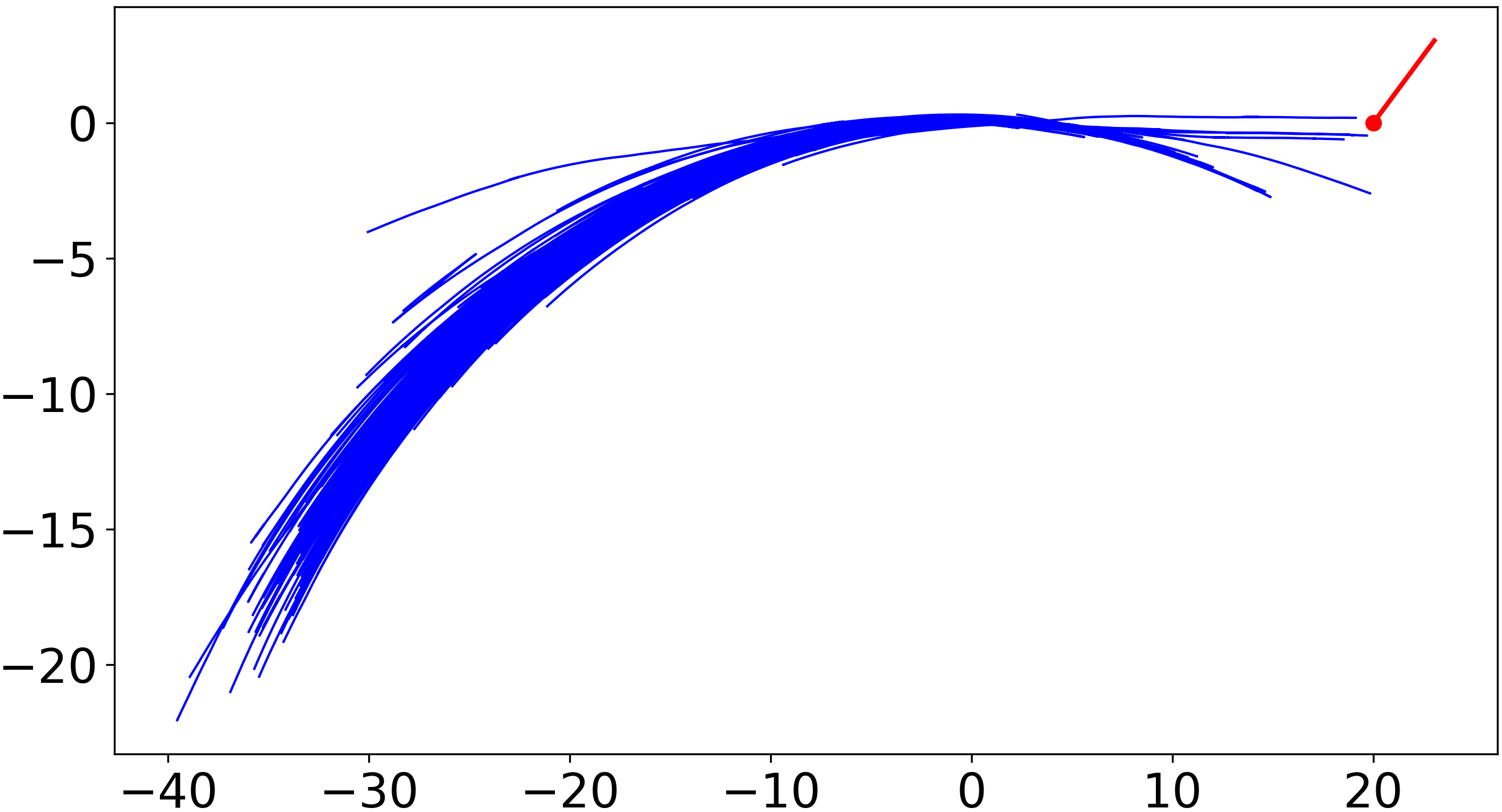}
\includegraphics[scale=0.06]{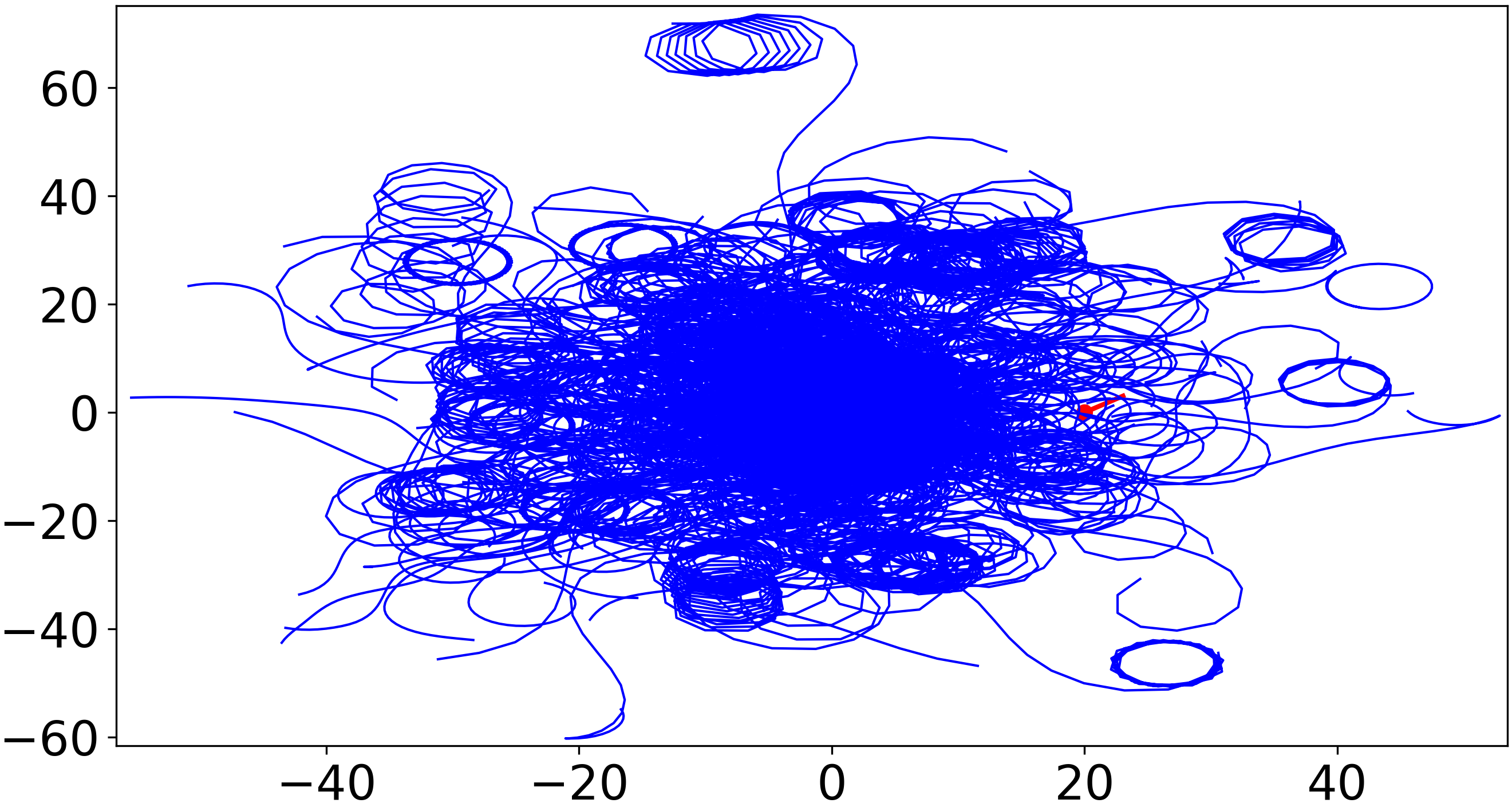}
\caption{Experiment 1. 1000 training trajectories resulting from the application DDPG (Left), PPO (Middle)  and TSHC (Right), respectively. The effect of virtual constraints on velocity is particularly visible for DDPG. For the given hyperparameter setting \cite[Tab. 3]{schulman2017proximal}, the trajectories for PPO have little spread and are favoring reverse driving. TSHC has a much better exploration strategy resulting from noise perturbations in the parameter space. The task is solved by TSHC in only 2.1s of learning time, when terminating upon the first solution found (no refinement step, no additional restart).}
\label{fig_ex1_ddpgppotshc}
\vspace{-0.5cm}
\end{figure*}

This paper is originally inspired by and most closely related to \cite{salimans2017evolution}. The main differences are discussed. The latter evolutionary (population-based) strategy updates parameters using a \emph{stochastic gradient estimate}. Thus, it updates $\theta \leftarrow \theta + \alpha \frac{1}{n \sigma}\sum_{i=1}^n R_i \zeta_i$, where hyperparameters $\alpha$ and $\sigma$ denote the learning rate and noise standard deviation, and where $R_i$ here indicates the stochastic scalar return provided by the simulation environment. This weighted averaging approach for the stochastic gradient estimate is not suitable for our control design method when using separate deterministic training tasks in combination with maximally sparse rewards. Here, hill climbing is more appropriate. This is since most of the $n$ trajectory candidates do not end up at $z^\text{goal}$ and are therefore not useful. Note also that only the introduction of virtual velocity constraints permitted us to quickly train with maximally sparse rewards. It is well known that for gradient-based training, especially of RNNs, the learning rate ($\alpha$ in \cite{salimans2017evolution}) is a critical hyperparameter choice. In the hill climbing setting this issue does not occur. Likewise, \emph{fitness shaping} \cite{wierstra2014natural}, also used in \cite{salimans2017evolution}, is not required. Note that above $\sigma$ has the same role as $\sigma_\text{pert}$. Except, in our setting, it additionally is adaptive according to Steps 29 and 31 in Algorithm \ref{alg_TSHC}. As implemented, this is only possible when training on multiple separate tasks. Other differences include the parallelization method in \cite{salimans2017evolution}, where random seeds shared among workers permit each worker to only need to send and receive the scalar return of an episode to and from each other worker. All perturbations and parameters are then reconstructed locally by each worker. Thus, for $n$ workers there are $n$ reconstructions at each parameter-iteration step. This requires precise control of each worker and can in rare cases lead to differing CPU utilizations among workers due to differing episode lengths. Therefore, they use a capping strategy on maximal episode length. In contrast, our proposed method is less sophisticated with one synchronized parameter update, which is then sent to all workers.

\section{Numerical Simulations\label{sec_expts}}

This section highlights different aspects of Algorithm \ref{alg_TSHC}. Numerical simulations of Sect. \ref{subsec_expt1} and \ref{subsec_expt2} were conducted on a laptop with an Intel Core i7 CPU @2.80GHz$\times$8, 15.6GB of memory, and with the only libraries employed Python's \texttt{numpy} and \texttt{multiprocessing}. Furthermore, in Sect. \ref{subsec_expt1} for the implementation of 2 comparative policy gradient methods, Tensorflow (without GPU-support) was used. Using these (for deep learning) very limited ressources enabled to evaluate the method's potential when significant computational power is not available. For more complex problems the latter is a necessity. Therefore, in Sect. \ref{subsec_expt3} TSHC is implemented in \texttt{Cuda C++} and 1 GPU is used.

\subsection{Experiment 1: Comparison with policy gradient methods\label{subsec_expt1}}

\begin{table}
\begin{center}
\vspace{0.2cm}
\caption{Experiment 1. Number of scalar parameters (weights) that need to be identified for DDPG, PPO and TSHC, respectively. TSHC requires to identify the least by a large margin, roughly by a factor 4. The fact that PPO here requires exactly four times the number of parameters of TSHC is a special case for $n_a=2$ controls (not generalizable for arbitrary $n_a$).} 
\label{tab_ex1_Nvar}
\bgroup
\def\arraystretch{1}
\begin{tabular}{|c|c|c|c|}
\hline
\rowcolor[gray]{0.8}   &  DDPG & PPO & TSHC \\ \hline 
$N_\text{var}$ &  19078 & 18440 &  4610 \\ 
\hline
\end{tabular}
\egroup
\end{center}
\vspace{-0.7cm}
\end{table}

To underline conceptual differences between TSHC and  2 policy gradient methods DDPG \cite{lillicrap2015continuous} and PPO \cite{schulman2017proximal}, a freeform navigation task with $z_0=[0,0,0,0]$ and $z^\text{goal}=[20,0,\pi/4,0]$ was considered, where vector $z=[x,~y,~\psi,~v]$ summarizes four of the vehicle's states. The same network architecture from \cite{schulman2017proximal} is used: a fully-connected MLP with 2 hidden layers of 64 units before the output layer. Eventhough this is the basic setup, considerable differences between DDPG, PPO and TSHC are implied. Both DDPG and PPO are each composed of a total of four networks: one actor, one critic, one actor target and one critic target network. For DDPG, further parameters result from batch normalization \cite{ioffe2015batch}. The number of parameters $N_\text{var}$ that need to be identified are indicated in Table \ref{tab_ex1_Nvar}. To enable a fair comparison, all of DPPG, PPO and TSHC are permitted to train on 1000 full rollouts according to their methods, whereby each rollout lasts at most $100$ timesteps. Thus, for TSHC, $N_\text{restarts}=1$ and $n=1000$ are set. For both PPO and DDPG, this implies 1000 iterations. Results are summarized in Fig. \ref{fig_ex1_ddpgppotshc}. The following observations can be made. First, in comparison to TSHC, for both DDPG and PPO significantly more parameters need to be identified, see Table \ref{tab_ex1_Nvar}. Second, DDPG and PPO do \emph{not} solve the task based on 1000 training simulations. In contrast, as Fig. \ref{fig_ex1_ddpgppotshc} demonstrates, TSHC has a much better exploration strategy resulting from noise perturbations in the \emph{parameter space}. It solves the task in just 2.1s. Finally, note that no $\sigma_\text{pert}$-iteration is conducted. It is not applicable since a single task is solved with an initial $\sigma_\text{pert}^\text{max}=10$. Because of these findings (other target poses were tested with qualitatively equivalent results) and the discussion in Sect. \ref{subsec_relatedWork_TSHC} about the handling of \emph{sparse rewards} and the fact that DDPG and PPO have no useful gradient direction for their parameter update or may average these out through random minibatch sampling, the focus in the subsequent sections is on TSHC and its analysis.


\subsection{Experiment 2: Inverted Pendulum\label{subsec_expt2}}

The discussion of tolerance levels in Sect. \ref{subsec_role_tolerances_eps} motivated to consider an alternative approach for tasks requiring stabilization. An analogy to optimal control is drawn. In linear finite horizon MPC,  closed-loop stability can be guaranteed through a terminal state constraint set which is invariant for a \emph{terminal controller}, often a linear quadratic regulator (LQR), see \cite{mayne2000constrained}. In a RL setting, the following procedure was considered. First, design a LQR for stabilization. Second, compute the region of attraction of the LQR controller \cite[Sect. 3.1.1]{tedrake2010lqr}. Third, use this region of attraction as stopping criterion, \emph{replacing} the heuristic $\epsilon$-tolerance selection. 

For evaluation, the inverted pendulum system equations and parameters from \cite{anderson1989learning} were adopted (four states, one input). However, in contrast to \cite{anderson1989learning}, which assumes just 2 discrete actions (maximum and minimum actuation force), here a continuous control variable is assumed which is limited by the 2 bounds, respectively. There are 2 basic problems: stabilization in the upright position with initial state in the same position, as well as a swing-up from the hanging position plus consequent stabilization in the upright position. For the application of TSHC, $(N_\text{restarts},N_\text{iter}^\text{max},n,T^\text{max},\beta,\sigma_\text{pert}^\text{max})=(3,100,100,500,2,10)$ are set, and the same MLP-architecture from Sect. \ref{subsec_expt1} is used. The following remarks can be made. First, the swing-up plus stabilization task was solved in $43.5$s runtime of TSHC (without refinement step) and using sparse rewards (obtained in the upright position $\pm 12^\circ$). For all three restarts a valid solution was generated. Note that $T^\text{max}=500$ in combination with a sampling time~\cite{anderson1989learning} of 0.02s corresponds to 10s simulation time. Stabilization in the upright position was achieved from 2.9s on. Rich reward signals were also tested, exploiting the deviation from current to goal angle as measure. However, rich rewards did not accelerate learning. 

In a second experiment, the objective was to \emph{simultaneously} encode the following 2 tasks in the network: stabilization in the upright position with initial state in the same position \emph{and} a swing-up from the hanging position plus consequent stabilization in the upright position. The runtime of TSHC (without refinement step) was $264.4$s, with 2 of 3 restarts returning a valid solution and using sparse rewards. Instead of learning both tasks simultaneously according to TSHC, it was also attempted to learn them by selecting one of the 2 tasks at random at every $i_\text{iter}$, and consequently conducting Step 6-41. Since the 2 tasks are quite different, this procedure could not encode a solution for \emph{both} tasks. This is mentioned to exemplify the importance of training \emph{simultaneously} on separate tasks, rather than training on a single tasks with  $(z_0,z^\text{goal})$-combinations varying over $i_\text{iter}$.

Finally, for the system parameters from \cite{anderson1989learning}, it was observed that the continuous control signal was operating mostly at saturated actuation bounds (switching in-between). This is mentioned for 2 reasons. First, aforementioned LQR-strategy could therefore never be applied since LQR assumes absence of state and input constraints. Second, it exemplifies the ease of RL-workflow with TSHC for quick nonlinear control design, even without significant system insights.

\begin{figure}
\vspace{0.3cm}
\centering%
\includegraphics[scale=1.0]{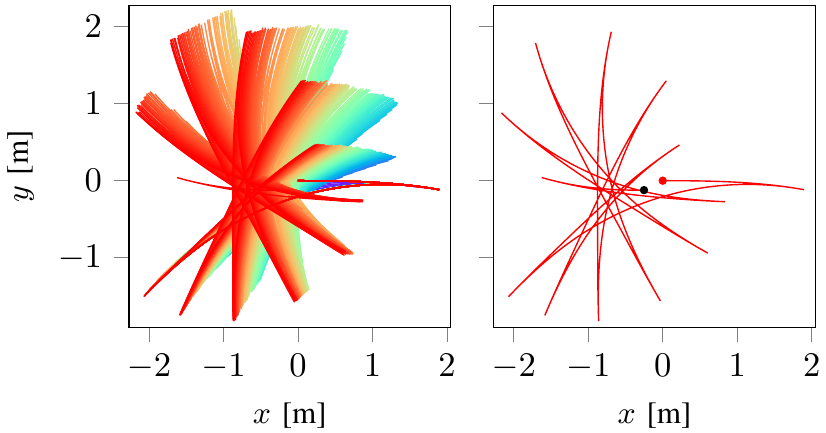}
\caption{Experiment 3. (Left) Display of \emph{all} 181 trajectories learnt and encoded in one MLP-[5,8,2]. Trajectories for each task are visualized in separate colors. (Right) Display of learnt result for only the most complex of the 181 training tasks, i.e., for $\psi^\text{goal}=180^\circ$. Recall that $\epsilon_\psi=1^\circ$ and $\epsilon_d=0.25$m. The vehicle's start and end CoG-position is indicated by red and black balls, respectively. As indicated, the transition involves frequent forward and backward driving but is constrained locally around the $(x,y)$-origin. See Sect. \ref{subsec_expt3} for further discussion.}
\label{fig:ex3}
\vspace{-0.6cm}
\end{figure}

\subsection{Experiment 3 and 4: GPU-based training\label{subsec_expt3}}

Experiment 3 is characterized by transitioning from $z_0=[0,0,0,0]$ to  $z^\text{goal}=[0,0,\psi^\text{goal},0]$ with $\psi^\text{goal}\in\{0,1,\dots,180\}$ measured in $[^\circ]$. This implies $N_\text{tasks}=181$. The feature vector is selected as $s_t=[\frac{x^\text{goal}-x_t}{\Delta x_n},\frac{y^\text{goal}-y_t}{\Delta y_n},\frac{\psi^\text{goal}-\psi_t}{\Delta \psi_n},\frac{v^\text{goal}-v_t}{\Delta v_n},a_t[0]]$ with normalization constants in the denominators and $a_t[0]\in[-1,1]$ indicating the steering angle-related network output (before scaling to $\delta_t$). A high-resolution tolerance of $\epsilon_\psi=1^\circ$ is set. In addition, $\epsilon_d=0.25$m and $\epsilon_v=5$km/h. Sampling time is 0.01s. As neural network, a MLP-[5,8,2] is used, which implies 1 hidden layer with 8 units. For selections $N_\text{restarts}=10$ and $N_\text{iter}^\text{max}=20$, MLP-[5,8,2] was the smallest possible network found to simultaneously encode all $181$ training tasks. The second variant of VVCs discussed in Sect. \ref{subsec_rt} is employed. Furthermore, $\sigma_\text{pert}\sim\mathcal{U}[10,1000]$, i.e., uniformly distributed at every $(i_\text{restart},i_\text{iter})$-combination.

\begin{figure}
\vspace{0.3cm}
\centering%
\includegraphics[scale=1.0]{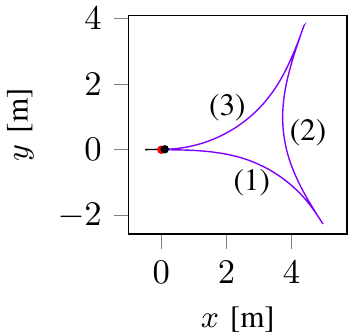}
\caption{Experiment 4. Display of learnt trajectory when encoding just \emph{one} task in a MLP-[5,8,2] for the transition from $\psi_0=0^\circ$ to $\psi^\text{goal}=180^\circ$. Brackets (1) and (3) imply forward, (2) reverse driving and their sequence. A black indicator visualizes the vehicle's final heading. Total learning time (including 10 restarts) was 10.6s. See Sect. \ref{subsec_expt3} for further discussion.}
\label{fig:ex4}
\vspace{-0.6cm}
\end{figure}

Several comments can be made. First, by application of \emph{control mirroring} w.r.t. steering the trained network enables to reach also  all of $\psi^\text{goal}\in\{181,\dots,360\}$. Second, the total learning time (runtime of TSHC) to encode all 181 training tasks was 31.1min. MLP-[5,8,2] implies a total of 66 parameters to learn. It is remarkable that such a small network has enough function approximation capability to encode all 181 tasks within limited training time and $\epsilon_\psi=1^\circ$. Third, note that the TSHC-trained network controller permits repeatable precision. As mentioned in of Sect. \ref{subsec_relatedWork_TSHC}, this is not attainable for stochastic policy gradient-based algorithms, which \emph{draw} their control signals, typically from a Gaussian distribution. Fourth, the learning results are visualized in Fig. \ref{fig:ex3}. These motivated to conduct an additional experiment with identical basic training setup (TSHC-settings, MLP-[5,8,2], etc.), however, now encoding only \emph{one} task for the transition from $\psi_0=0^\circ$ to $\psi^\text{goal}=180^\circ$. The result is visualized in Fig. \ref{fig:ex4}. Notice the much reduced number of switches between forward and backward driving, and the different $(x,y)$-range.

The comparison of  Fig. \ref{fig:ex3} and \ref{fig:ex4} emphasizes the interesting observation that the more motion primitives are encoded in a single network the less performant the single learnt motion trajectories are. This is believed to illustrate potential of partitioning the total number of designated tasks into \emph{subsets} of training tasks for which separate networks are then learnt using TSHC. The promised advantages include faster overall trainig times, higher performance of learnt trajectories, and ability to employ tiny networks with few parameters for each subset. Work in this perspective is subject of ongoing work.

\section{Conclusion\label{sec_conclusion}}

Within the context of automated vehicles, for the design of model-based controllers parameterized by neural networks a simple gradient-free reinforcement learning algorithm labeled TSHC was proposed. The concept of (i) training on separate tasks with the purpose of encoding motion primitives, and (ii) employing sparse rewards in combinations with virtual velocity constraints in setpoint proximity were specifically advocated. Aspects of TSHC were illustrated in 4 numerical experiments. The presented method is not limited to automated driving. Most real-world learning applications for control systems, especially in robotics, are characterized by sparse rewards and the availability of high-fidelity system models that can be leveraged for offline training. 

Subject of future work is focus on system models of various complexity (e.g., kinematic vs. dynamic vehicle models), the partitioninig of tasks into separate subsets of tasks for which separate network parametrizations are learnt, analysis of different feature vectors and closed-loop evaluation.

%

%
%
%
%
%
%

\nocite{*}
\bibliographystyle{ieeetr}
\bibliography{myref}

\begin{thebibliography}{10}

\bibitem{paden2016survey}
B.~Paden, M.~{\v{C}}{\'a}p, S.~Z. Yong, D.~Yershov, and E.~Frazzoli, ``A survey
  of motion planning and control techniques for self-driving urban vehicles,''
  {\em IEEE Transactions on Intelligent Vehicles}, vol.~1, no.~1, pp.~33--55,
  2016.

\bibitem{karaman2011anytime}
S.~Karaman, M.~R. Walter, A.~Perez, E.~Frazzoli, and S.~Teller, ``Anytime
  motion planning using the {RRT},'' in {\em IEEE Conference on Robotics and
  Automation}, pp.~1478--1483, 2011.

\bibitem{dolgov2010path}
D.~Dolgov, S.~Thrun, M.~Montemerlo, and J.~Diebel, ``Path planning for
  autonomous vehicles in unknown semi-structured environments,'' {\em The
  International Journal of Robotics Research}, vol.~29, no.~5, pp.~485--501,
  2010.

\bibitem{mcnaughton2011motion}
M.~McNaughton, C.~Urmson, J.~M. Dolan, and J.-W. Lee, ``Motion planning for
  autonomous driving with a conformal spatiotemporal lattice,'' in {\em IEEE
  Conference on Robotics and Automation}, pp.~4889--4895, 2011.

\bibitem{frazzoli1999hybrid}
E.~Frazzoli, M.~A. Dahleh, and E.~Feron, ``A hybrid control architecture for
  aggressive maneuvering of autonomous helicopters,'' in {\em IEEE Conference
  on Decision and Control}, vol.~3, pp.~2471--2476, 1999.

\bibitem{schouwenaars2003robust}
T.~Schouwenaars, B.~Mettler, E.~Feron, and J.~P. How, ``Robust motion planning
  using a maneuver automation with built-in uncertainties,'' in {\em IEEE
  American Control Conference}, vol.~3, pp.~2211--2216, 2003.

\bibitem{gray2012predictive}
A.~Gray, Y.~Gao, T.~Lin, J.~K. Hedrick, H.~E. Tseng, and F.~Borrelli,
  ``Predictive control for agile semi-autonomous ground vehicles using motion
  primitives,'' in {\em IEEE American Control Conference}, pp.~4239--4244,
  2012.

\bibitem{liniger2015optimization}
A.~Liniger, A.~Domahidi, and M.~Morari, ``Optimization-based autonomous racing
  of 1: 43 scale rc cars,'' {\em Optimal Control Applications and Methods},
  vol.~36, no.~5, pp.~628--647, 2015.

\bibitem{falcone2007predictive}
P.~Falcone, F.~Borrelli, J.~Asgari, H.~E. Tseng, and D.~Hrovat, ``Predictive
  active steering control for autonomous vehicle systems,'' {\em IEEE
  Transactions on Control Systems technology}, vol.~15, no.~3, pp.~566--580,
  2007.

\bibitem{plessen2017spatial}
M.~G. Plessen, D.~Bernardini, H.~Esen, and A.~Bemporad, ``Spatial-based
  predictive control and geometric corridor planning for adaptive cruise
  control coupled with obstacle avoidance,'' {\em IEEE Transactions on Control
  Systems Technology}, 2017.

\bibitem{plessen2017trajectory2}
M.~G. Plessen, ``Trajectory planning of automated vehicles in tube-like road
  segments,'' in {\em IEEE Conference on Intelligent Transportation Systems},
  pp.~83--88, 2017.

\bibitem{plessen2017trajectory}
M.~G. Plessen, P.~F. Lima, J.~M{\aa}rtensson, A.~Bemporad, and B.~Wahlberg,
  ``Trajectory planning under vehicle dimension constraints using sequential
  linear programming,'' in {\em IEEE Conference on Intelligent Transportation
  Systems}, pp.~108--113, 2017.

\bibitem{pomerleau1989alvinn}
D.~A. Pomerleau, ``{ALVINN}: An autonomous land vehicle in a neural network,''
  in {\em Advances in Neural Information Processing Systems}, pp.~305--313,
  1989.

\bibitem{bojarski2016end}
M.~Bojarski, D.~Del~Testa, D.~Dworakowski, B.~Firner, B.~Flepp, P.~Goyal, {\em
  et~al.}, ``End to end learning for self-driving cars,'' {\em arXiv preprint
  arXiv:1604.07316}, 2016.

\bibitem{chen2017brain}
S.~Chen, S.~Zhang, J.~Shang, B.~Chen, and N.~Zheng, ``Brain inspired cognitive
  model with attention for self-driving cars,'' {\em arXiv preprint
  arXiv:1702.05596}, 2017.

\bibitem{chen2015deepdriving}
C.~Chen, A.~Seff, A.~Kornhauser, and J.~Xiao, ``Deepdriving: Learning
  affordance for direct perception in autonomous driving,'' in {\em IEEE
  International Conference on Computer Vision}, pp.~2722--2730, 2015.

\bibitem{xu2016end}
H.~Xu, Y.~Gao, F.~Yu, and T.~Darrell, ``End-to-end learning of driving models
  from large-scale video datasets,'' {\em arXiv preprint arXiv:1612.01079},
  2016.

\bibitem{paxton2017combining}
C.~Paxton, V.~Raman, G.~D. Hager, and M.~Kobilarov, ``Combining neural networks
  and tree search for task and motion planning in challenging environments,''
  {\em arXiv preprint arXiv:1703.07887}, 2017.

\bibitem{urmson2008autonomous}
C.~Urmson, J.~Anhalt, D.~Bagnell, C.~Baker, R.~Bittner, M.~Clark, J.~Dolan,
  D.~Duggins, T.~Galatali, C.~Geyer, {\em et~al.}, ``Autonomous driving in
  urban environments: Boss and the urban challenge,'' {\em Journal of Field
  Robotics}, vol.~25, no.~8, pp.~425--466, 2008.

\bibitem{gillespie1997vehicle}
T.~D. Gillespie, ``Vehicle dynamics,'' {\em Warren dale}, 1997.

\bibitem{rajamani2011vehicle}
R.~Rajamani, {\em Vehicle dynamics and control}.
\newblock Springer Science \& Business Media, 2011.

\bibitem{glasmachers2017limits}
T.~Glasmachers, ``Limits of end-to-end learning,'' {\em arXiv preprint
  arXiv:1704.08305}, 2017.

\bibitem{national2014traffic}
{National Highway Traffic Safety Administration}, ``Traffic safety facts, 2014:
  a compilation of motor vehicle crash data from the fatality analysis
  reporting system and the general estimates system. dot hs 812261,'' {\em
  Department of Transportation, Washington, DC}, 2014.

\bibitem{plessen2016multi}
M.~G. Plessen, D.~Bernardini, H.~Esen, and A.~Bemporad, ``Multi-automated
  vehicle coordination using decoupled prioritized path planning for multi-lane
  one-and bi-directional traffic flow control,'' in {\em IEEE Conference on
  Decision and Control}, pp.~1582--1588, 2016.

\bibitem{salimans2017evolution}
T.~Salimans, J.~Ho, X.~Chen, and I.~Sutskever, ``Evolution strategies as a
  scalable alternative to reinforcement learning,'' {\em arXiv preprint
  arXiv:1703.03864}, 2017.

\bibitem{akiba2017extremely}
T.~Akiba, S.~Suzuki, and K.~Fukuda, ``Extremely large minibatch sgd: Training
  resnet-50 on imagenet in 15 minutes,'' {\em arXiv preprint arXiv:1711.04325},
  2017.

\bibitem{teslaP100}
{Nvidia}, ``Tesla {P}100.''
  \url{https://images.nvidia.com/content/tesla/pdf/nvidia-tesla-p100-PCIe-datasheet.pdf},
  2016.

\bibitem{gers2002learning}
F.~A. Gers, N.~N. Schraudolph, and J.~Schmidhuber, ``Learning precise timing
  with {LSTM} recurrent networks,'' {\em Journal of Machine Learning Research},
  vol.~3, no.~Aug, pp.~115--143, 2002.

\bibitem{cho2014learning}
K.~Cho, B.~Van~Merri{\"e}nboer, C.~Gulcehre, D.~Bahdanau, F.~Bougares,
  H.~Schwenk, and Y.~Bengio, ``Learning phrase representations using rnn
  encoder-decoder for statistical machine translation,'' {\em arXiv preprint
  arXiv:1406.1078}, 2014.

\bibitem{jozefowicz2015empirical}
R.~Jozefowicz, W.~Zaremba, and I.~Sutskever, ``An empirical exploration of
  recurrent network architectures,'' in {\em International Conference on
  Machine Learning}, pp.~2342--2350, 2015.

\bibitem{lillicrap2015continuous}
T.~P. Lillicrap, J.~J. Hunt, A.~Pritzel, N.~Heess, T.~Erez, Y.~Tassa,
  D.~Silver, and D.~Wierstra, ``Continuous control with deep reinforcement
  learning,'' {\em arXiv preprint arXiv:1509.02971}, 2015.

\bibitem{sutton1998reinforcement}
R.~S. Sutton and A.~G. Barto, {\em Reinforcement learning: An introduction},
  vol.~1.
\newblock MIT press Cambridge, 1998.

\bibitem{bengio2009curriculum}
Y.~Bengio, J.~Louradour, R.~Collobert, and J.~Weston, ``Curriculum learning,''
  in {\em International Conference on Machine Learning}, pp.~41--48, ACM, 2009.

\bibitem{randlov1998learning}
J.~Randlov and P.~Alstrom, ``Learning to drive a bicycle using reinforcement
  learning and shaping,'' in {\em International Conference on Machine
  Learning}, pp.~463--471, 1998.

\bibitem{koutnik2014online}
J.~Koutn{\'\i}k, J.~Schmidhuber, and F.~Gomez, ``Online evolution of deep
  convolutional network for vision-based reinforcement learning,'' in {\em
  International Conference on Simulation of Adaptive Behavior}, pp.~260--269,
  Springer, 2014.

\bibitem{heess2017emergence}
N.~Heess, S.~Sriram, J.~Lemmon, J.~Merel, G.~Wayne, Y.~Tassa, {\em et~al.},
  ``Emergence of locomotion behaviours in rich environments,'' {\em arXiv
  preprint arXiv:1707.02286}, 2017.

\bibitem{anderson1989learning}
C.~W. Anderson, ``Learning to control an inverted pendulum using neural
  networks,'' {\em IEEE Control Systems Magazine}, vol.~9, no.~3, pp.~31--37,
  1989.

\bibitem{siegelmann1991turing}
H.~T. Siegelmann and E.~D. Sontag, ``Turing computability with neural nets,''
  {\em Applied Mathematics Letters}, vol.~4, no.~6, pp.~77--80, 1991.

\bibitem{schulman2017proximal}
J.~Schulman, F.~Wolski, P.~Dhariwal, A.~Radford, and O.~Klimov, ``Proximal
  policy optimization algorithms,'' {\em arXiv preprint arXiv:1707.06347},
  2017.

\bibitem{fu2005simulation}
M.~C. Fu, F.~W. Glover, and J.~April, ``Simulation optimization: a review, new
  developments, and applications,'' in {\em IEEE Winter Simulation Conference},
  pp.~13--pp, IEEE, 2005.

\bibitem{wierstra2014natural}
D.~Wierstra, T.~Schaul, T.~Glasmachers, Y.~Sun, J.~Peters, and J.~Schmidhuber,
  ``Natural evolution strategies.,'' {\em Journal of Machine Learning
  Research}, vol.~15, no.~1, pp.~949--980, 2014.

\bibitem{xu2010industrial}
J.~Xu, B.~L. Nelson, and J.~Hong, ``Industrial strength {COMPASS}: {A}
  comprehensive algorithm and software for optimization via simulation,'' {\em
  ACM Transactions on Modeling and Computer Simulation}, vol.~20, no.~1, p.~3,
  2010.

\bibitem{hong2009brief}
L.~J. Hong and B.~L. Nelson, ``A brief introduction to optimization via
  simulation,'' in {\em IEEE Winter Simulation Conference}, pp.~75--85, 2009.

\bibitem{sutton2000policy}
R.~S. Sutton, D.~A. McAllester, S.~P. Singh, and Y.~Mansour, ``Policy gradient
  methods for reinforcement learning with function approximation,'' in {\em
  Advances in Neural Information Processing Systems}, pp.~1057--1063, 2000.

\bibitem{mnih2016asynchronous}
V.~Mnih, A.~P. Badia, M.~Mirza, A.~Graves, T.~Lillicrap, T.~Harley, D.~Silver,
  and K.~Kavukcuoglu, ``Asynchronous methods for deep reinforcement learning,''
  in {\em International Conference on Machine Learning}, pp.~1928--1937, 2016.

\bibitem{silver2014deterministic}
D.~Silver, G.~Lever, N.~Heess, T.~Degris, D.~Wierstra, and M.~Riedmiller,
  ``Deterministic policy gradient algorithms,'' in {\em International
  Conference on Machine Learning}, pp.~387--395, 2014.

\bibitem{geering2011stochastic}
H.~Geering, G.~Dondi, F.~Herzog, and S.~Keel, ``Stochastic systems,'' {\em
  Course script}, 2011.

\bibitem{ioffe2015batch}
S.~Ioffe and C.~Szegedy, ``Batch normalization: Accelerating deep network
  training by reducing internal covariate shift,'' in {\em International
  Conference on Machine Learning}, pp.~448--456, 2015.

\bibitem{ba2016layer}
J.~L. Ba, J.~R. Kiros, and G.~E. Hinton, ``Layer normalization,'' {\em arXiv
  preprint arXiv:1607.06450}, 2016.

\bibitem{mayne2000constrained}
D.~Q. Mayne, J.~B. Rawlings, C.~V. Rao, and P.~O. Scokaert, ``Constrained model
  predictive control: Stability and optimality,'' {\em Automatica}, vol.~36,
  no.~6, pp.~789--814, 2000.

\bibitem{tedrake2010lqr}
R.~Tedrake, I.~R. Manchester, M.~Tobenkin, and J.~W. Roberts, ``{LQR}-trees:
  Feedback motion planning via sums-of-squares verification,'' {\em
  International Journal of Robotics Research}, vol.~29, no.~8, pp.~1038--1052,
  2010.

\bibitem{long2015fully}
J.~Long, E.~Shelhamer, and T.~Darrell, ``Fully convolutional networks for
  semantic segmentation,'' in {\em IEEE Conference on Computer Vision and
  Pattern Recognition}, pp.~3431--3440, 2015.

\end{thebibliography}


\end{document}